\pdfoutput=1
\documentclass[11pt]{article}

\usepackage[final]{emnlp2021}
\usepackage{times}
\usepackage{latexsym}

\usepackage{amsmath,amsfonts,bm}

\def\eqref#1{equation~\ref{#1}}
\def\1{\bm{1}}

\def\rmM{{\mathbf{M}}}

\def\rmS{{\mathbf{S}}}

\def\rmW{{\mathbf{W}}}

\DeclareMathAlphabet{\mathsfit}{\encodingdefault}{\sfdefault}{m}{sl}
\SetMathAlphabet{\mathsfit}{bold}{\encodingdefault}{\sfdefault}{bx}{n}

\def\sR{{\mathbb{R}}}

\newcommand\fref{Fig.~\ref}
\newcommand\tref{Tab.~\ref}

\usepackage{hyperref}
\usepackage{url}
\usepackage{graphicx,subcaption}
\usepackage{booktabs}
\usepackage{multirow}

\newcommand\blfootnote[1]{%
  \begingroup
  \renewcommand\thefootnote{}\footnote{#1}%
  \addtocounter{footnote}{-1}%
  \endgroup
}

\title{What’s Hidden in a One-layer Randomly Weighted Transformer?}

\author{
Sheng Shen$^{\dagger*}$, Zhewei Yao$^{\dagger*}$,  Douwe Kiela$^\ddagger$, Kurt Keutzer$^\dagger$ and Michael W. Mahoney$^\dagger$\\
 $^\dagger$UC Berkeley; $^\ddagger$Facebook AI Research\\
 \texttt{\{sheng.s,zheweiy\}@berkeley.edu, dkiela@fb.com}
}

\begin{document}
\maketitle

%%%%%%%% BODY TEXT
\begin{abstract}
We demonstrate that, hidden within one-layer randomly weighted neural networks, there exist subnetworks that can achieve impressive performance, without ever modifying the weight initializations, on machine translation tasks. 
To find subnetworks for one-layer randomly weighted neural networks, we apply different binary masks to the same weight matrix to generate different layers. 
Hidden within a one-layer randomly weighted Transformer, we find that subnetworks that can achieve 29.45/17.29 BLEU on IWSLT14/WMT14. 
Using a fixed pre-trained embedding layer, the previously found subnetworks are smaller than, but can match 98\%/92\% (34.14/25.24 BLEU) of the performance of, a trained Transformer$_\text{small/base}$ on IWSLT14/WMT14. 
Furthermore, we demonstrate the effectiveness of larger and deeper transformers in this setting, as well as the impact of different initialization methods.\footnote{We released the source code at \url{https://github.com/sIncerass/one_layer_lottery_ticket}.}
\end{abstract}
\blfootnote{$^*$Equal contribution.}
% \vspace{-4mm}
\vspace{-2mm}
\section{Introduction}
\label{sec:intro}
\vspace{-1mm}

Modern deep learning often trains millions or even billions of parameters~\citep{Devlin:2018bert,shoeybi2019megatron,raffel2019exploring,brown2020language} to deliver good performance for a model. 
Recently, \citet{frankle2018lottery,frankle2020linear} demonstrated that these over-parameterized  networks contain sparse subnetworks, when trained in isolation, that can achieve similar or better performance than the original~model.

Furthermore, recent studies revisit the initialization stage of finding these subnetworks in vision models~\citep{Zhou:2019deconstructing,Ramanujan:2020hidden}.
Such a mask, which is used to mask out a part of the entire network to those subnetworks, is referred to as a ``Supermask.'' 
That is to say, subnetworks of a \textit{randomly weighted neural network} (NN) can achieve competitive performance, which may act as a good  ``prior''~\citep{gaier2019weight} and connect to the long history of leveraging random features~\citep{Gamba:1961papa,Baum:1988jc} and/or random kernel methods~\citep{Rahimi:2008random,Rahimi:2009kitchen} in machine learning. 
Here, we examine the following question: how does a fully randomized natural language processing (NLP) model perform in the multi-layer setting, and particularly in the (so far under-explored) one-layer setting?

\begin{figure}
    \centering
    \includegraphics[width=0.8\linewidth]{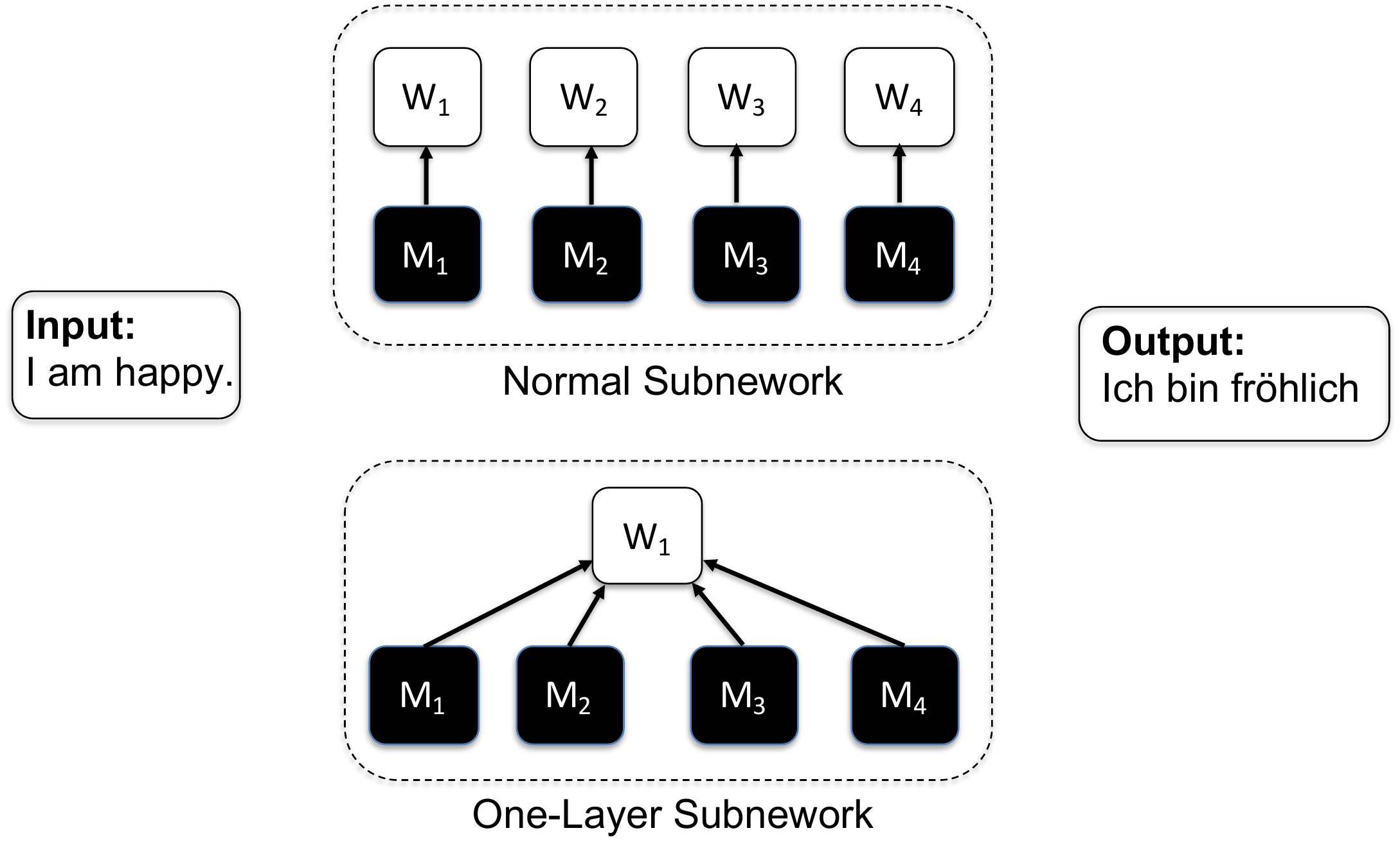}
            \vspace{-5pt}
    \caption{ Illustration plot for a normal subnetwork and a one-layer subnetwork.}
    \label{fig:model_illustration}
    \vspace{-6mm}
\end{figure}

In this work, we first validate that there exist subnetworks of standard randomly weighted Transformers (Reservoir Transformers in~\citep{shen2020reservoir}) that can perform competitively with fully-weighted alternatives on machine translation and natural language understanding tasks. 
With 50\% randomized weights remaining, we found a subnetwork that can reach 29.45/17.29 BLEU on IWSLT14/WMT14, respectively. 
We also investigate the special case of finding subnetworks in one-layer randomly weighted Transformers (see~\fref{fig:model_illustration}). 
To obtain the subnetworks, we repeatedly apply the same randomized Transformer layer several times with different Supermasks. 
The resulting subnetwork of a one-layer randomly-weighted Transformer has similar performance as the multi-layer counterparts with a 30\% lower memory footprint. 
We also study the impact of different depths/widths of Transformers along with the effectiveness of two initialization methods. 
Finally, using the pre-trained embedding layers, we find that the subnetworks hidden in one layer randomly weighted Transformer$_\text{wide/wider}$ are smaller than, but can match 98\%/92\% of the performance of, a trained Transformer$_\text{small/base}$ on IWSLT14/WMT14. 
We hope our findings can offer new insights for understanding Transformers. 
% \vspace{-4mm}
% \vspace{-3mm}
\section{Related Work}
\vspace{-1mm}
\noindent\textbf{Lottery Tickets Hypothesis.}
\citet{frankle2018lottery} found that NNs for computer vision contain subnetworks that can be effectively trained from scratch when reset to their initialization.
Subsequent works~\citep{Zhou:2019deconstructing,Ramanujan:2020hidden,wortsman2020supermasks} demonstrated that so-called winning tickets can achieve performance without training, where the mask for finding the subnetwork at initialization is called ``supermask.'' 
In NLP, previous works find that matching subnetworks exist early in training with Transformers~\citep{yu2019playing}, LSTMs~\citep{renda2020comparing}, and fully-weighted per-trained BERT~\citep{chen2020lottery,prasanna2020bert} or Vison-and-Language model~\citep{gan2021playing}, but not at \textit{initialization}. 

\noindent\textbf{Random Feature.} 
In the early days of neural networks, fixed random layers~\citep{Baum:1988jc,Schmidt:1992pr,Pao:1994nc} have been studied in reservoir computing~\citep{Maass:2002lsm,Jaeger:2003echostate,Lukovsevivcius:2009reservoir}, ``random kitchen sink'' kernel machines~\citep{Rahimi:2008random,Rahimi:2009kitchen}, and so on. 
Recently, random features have also been extensively explored for modern neural networks in deep reservoir computing networks \citep{Scardapane:2017randomness,Gallicchio:2017echo,shen2020reservoir}, random kernel feature~\citep{peng2021random,Choromanski:2020performer}, and  applications in text classification~\citep{Conneau:2017infersent,Wieting:2019notraining}, summarization \citep{Pilault:2020impressive} and probing \citep{voita2020information}. 

\noindent\textbf{Compressing Transformer.} A wide range of neural network compression techniques have been applied to Transformers. 
This includes pruning  \citep{fan2019reducing,michel2019sixteen,sanh2020movement,yao2021mlpruning} where parts of the model weights are dropped, parameter-sharing \citep{lan2020albert,dehghani2018universal,bai2019deep} where the same
parameters are used in different parts of a model, quantization \citep{shen2020q,li2020train}
where the weights of the Transformer model are represented with fewer bits, and distilliation \citep{sun2020mobilebert,jiao2020tinybert} where a compact student model is trained to mimic a larger teacher model. 
To find the proposed subnetwork at initialization, we develop our method in the spirit of parameter sharing and pruning. 
\vspace{-2mm}
\section{Methodology}
\label{sec:methodology}
\vspace{-1mm}

\noindent\textbf{Finding a Supermask for Randomly Weighted Transformer.} 
In a general pruning framework, denote weight matrix as $\rmW\in\sR^{d\times d}$  ($\rmW$ could be a non-square matrix), input as $x\in\sR^d$ and the network as $f(x; \rmW)$. 
A subnetwork defined is $f(x; \rmW \odot \rmM)$, where $\rmM\in\sR^{d\times d}$ is a binary matrix and $\odot$ is the element-wise product. 
To find the subnetwork for a randomly weighted network, $\rmM\in\sR^{d\times d}$ is trained while $\rmW$ is kept at a random initialization. 
Following~\citet{Ramanujan:2020hidden}, denote $\rmS\in\sR^{d\times d}$ as the associated importance score matrix of $\rmW$, which is learnable during training. 
We keep top-k percents of weights by the importance score of $\rmS$ to compute $\rmM$, i.e., 
\begin{equation*}
\label{eq:topk}
\resizebox{\linewidth}{!}{$\displaystyle
% \small
\rmM = \text{Top}_k(\rmS), \text{where}
    ~\text{Top}_k(\rmS_{i, j}) = 
    \begin{cases}
    ~1 &\rmS_{i, j}\text{ in top k\%},\\
    ~0 &\text{else}.
    \end{cases}
    $
}
\end{equation*}
Note that $\text{Top}_k$ is an undifferentiated function.
To enable training of $\rmS$, we use the straight-through gradient estimator \citep{bengio2013estimating}, in which $\text{Top}_k$ is treated as the identity in backpropagation. 
During inference, we can simply construct and store the binary Supermask $\rmM$ and the floating-point $\rmW$ while dropping $\rmS$ for future usage. 

\noindent\textbf{One-layer randomly weighted Transformer.}
We use the Transformer architecture (see ~\newcite{Vaswani:2017attention} for more details).
For a general randomly weighted Transformer model with Supermask, there exist $\rmM_l$s and $\rmW_l$s for all layers $l \in \{1, ... L\}$. 
Due to the natural property of layer stacking in Transformers, all $\rmW_l$s have the same shape with the same initialization method. 
This leads to an unexplored question: ``What's hidden in a one-layer (instead of L-layer) randomly weighted transformer?''

\begin{figure*}
    \centering
    \begin{subfigure}{.4\textwidth}
        \centering
        \includegraphics[width=\textwidth]{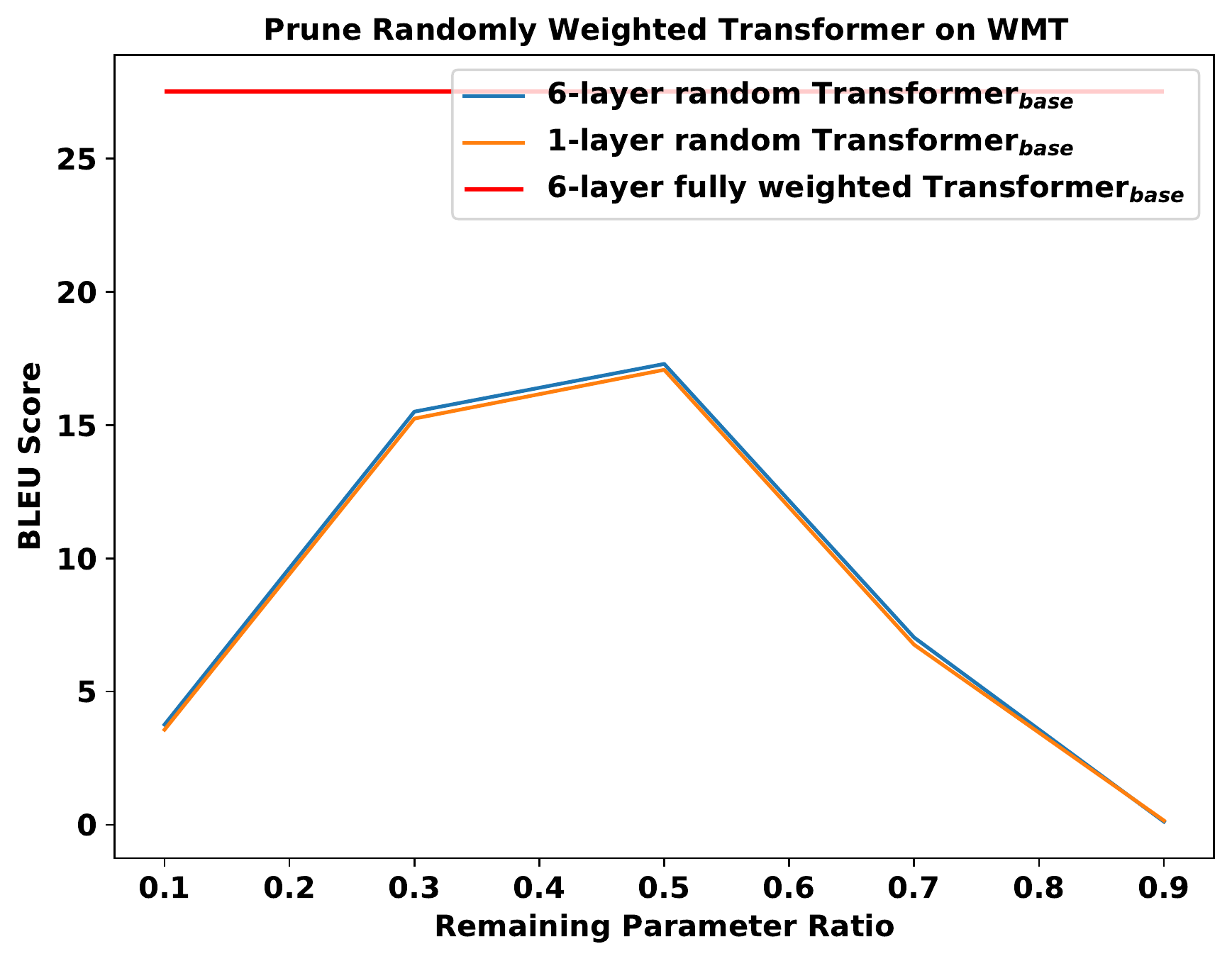}
    \end{subfigure}%
    \begin{subfigure}{.4\textwidth}
        \centering
        \includegraphics[width=\textwidth]{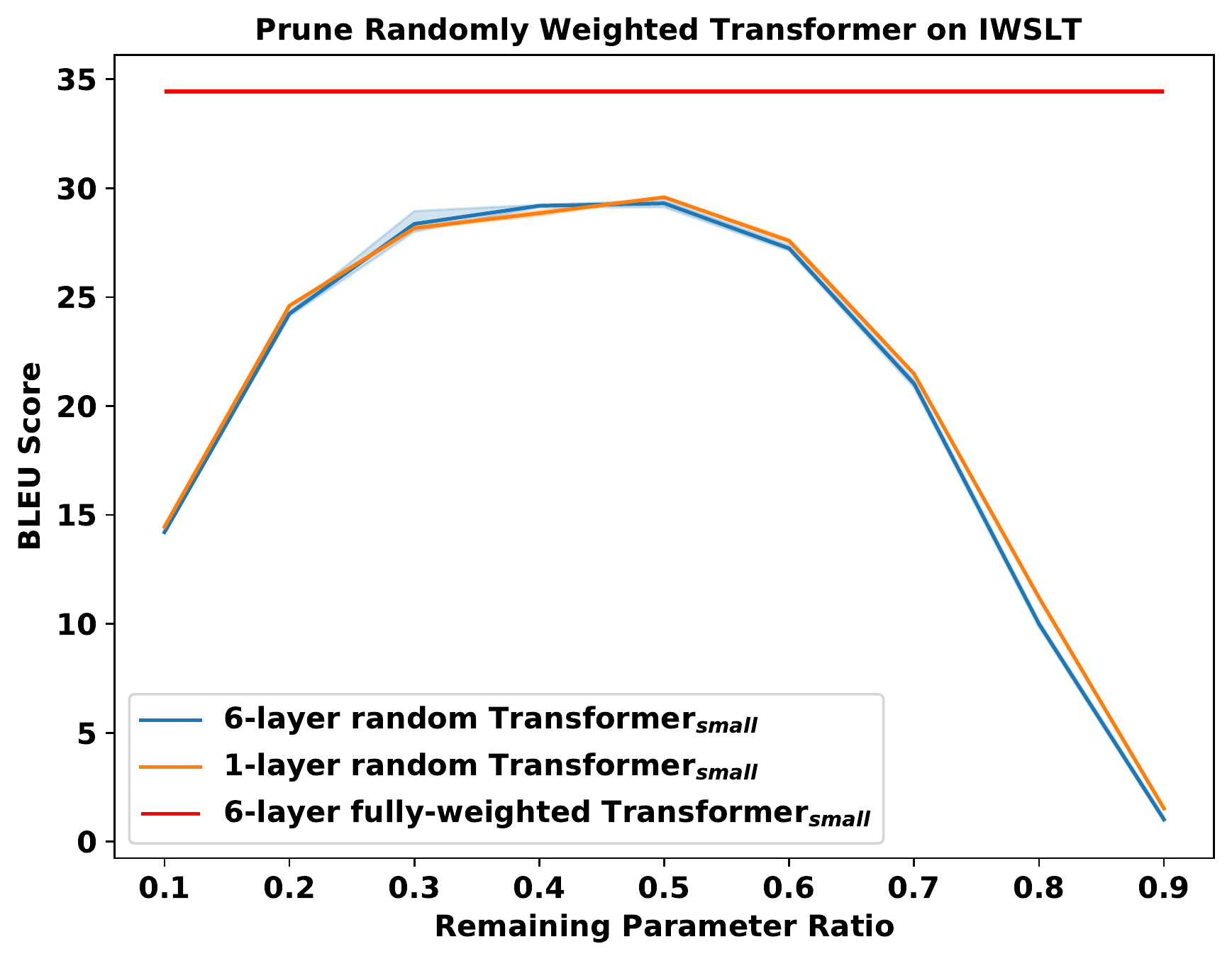}
    \end{subfigure}
            \vspace{-5pt}
    \caption{Prune Randomly Weighted Transformer performance on WMT14 (left) and IWSLT14 (right).}
    \label{fig:mt_base}
            \vspace{-5pt}
\end{figure*}

\begin{figure*}
    \centering
    \begin{subfigure}{.4\textwidth}
        \centering
        \includegraphics[width=\textwidth]{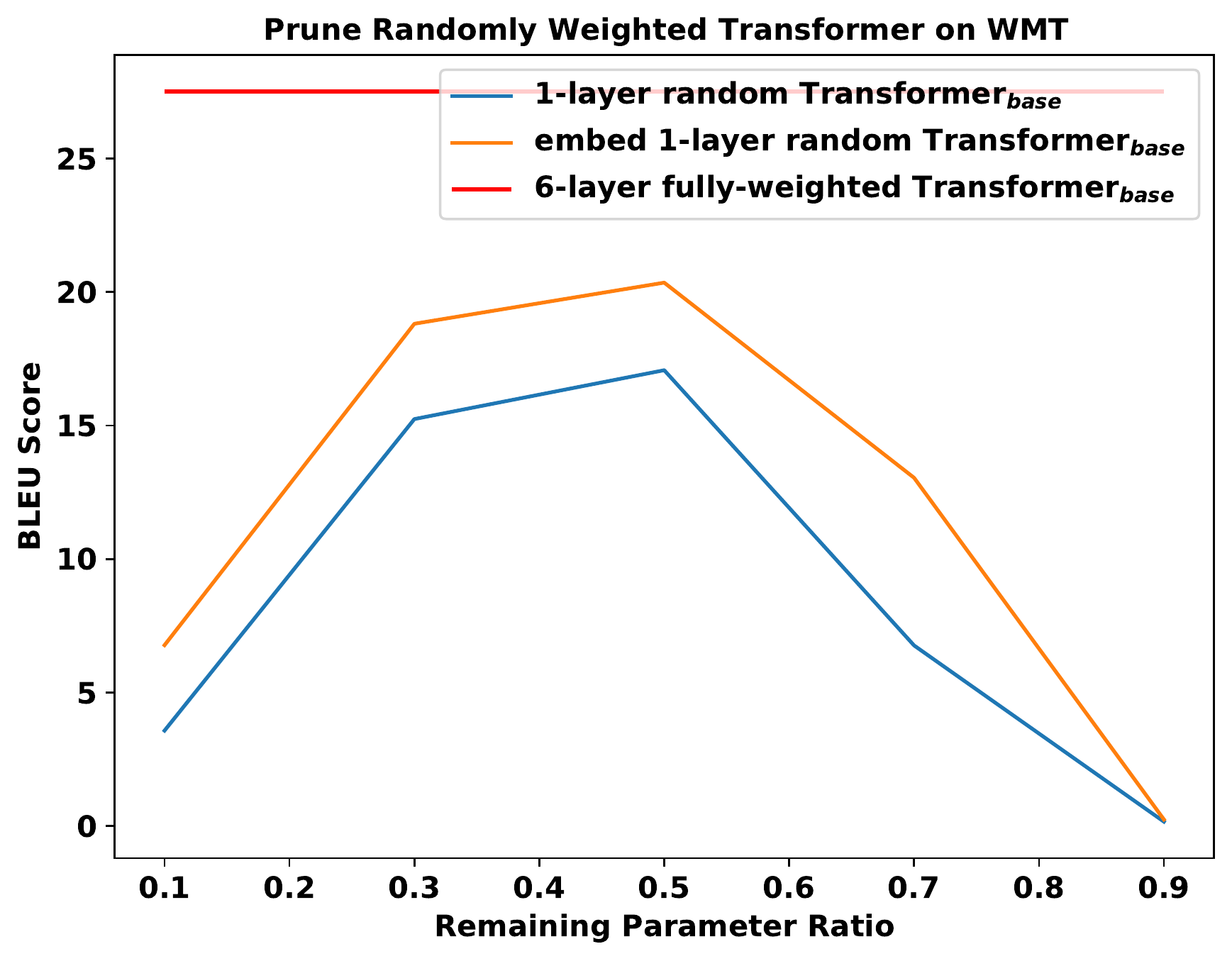}
    \end{subfigure}%
    \begin{subfigure}{.4\textwidth}
        \centering
        \includegraphics[width=\textwidth]{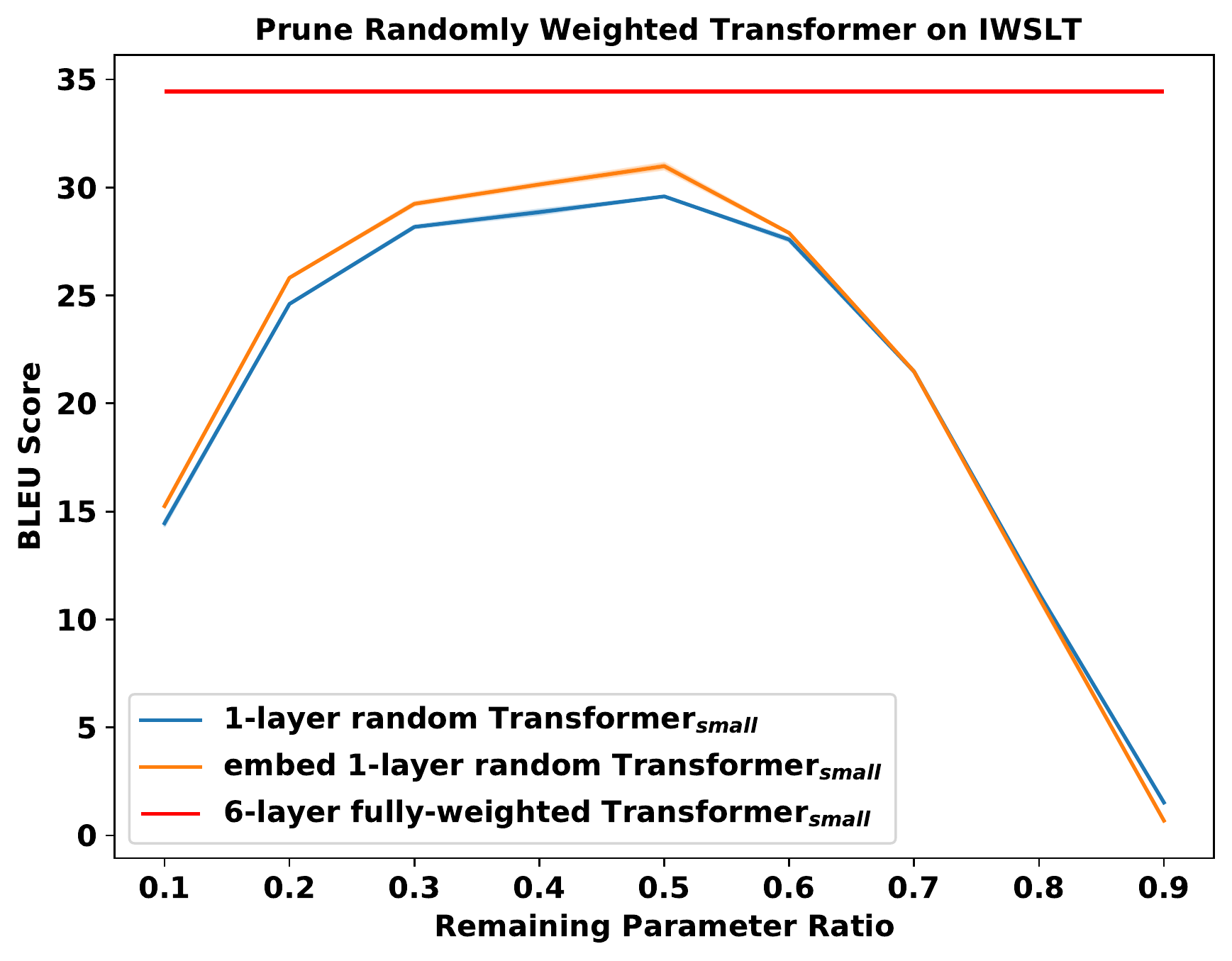}
    \end{subfigure}
            \vspace{-5pt}
    \caption{The effectiveness of pre-trained embedding layers on WMT14 (left) and IWSLT14 (right).}
    \label{fig:mt_embed}
            \vspace{-10pt}

\end{figure*}

Let us use a toy example to explain why there is no need for $L$ redundant $\rmW_l$s. 
Assume that, for a random weighted matrix $\rmW_l$, the probability that it has a ``good'' subnetwork is $p$\footnote{Here, the ``good'' can be any defined metric, e.g., $(\rmM\odot\rmW_l)\text{x}\approx\rmW^*\text{x}$ for all $x$ and a pre-defined $\rmW^*$.}. 
Furthermore, assume that for two different layers, the probability that both have the ``good'' subnetworks is independent.
Then for $L$ different layers, the probability that all $\rmW_l$s have the ``good'' subnetworks is $p^L$. 
Meanwhile, since $\rmW_1$ has the same initialization method as $\rmW_l$, the probability that $\rmW_1$ has a ``good'' subnetwork for $l$-th layer is also $p$.
Thus, for $L$ different layers, the probability that using $\rmW_1$ to generate all ``good'' subnetworks is also $p^L$. 

In this paper, we investigate the scenario where one randomized layer is applied for $L$ times repeatedly with $L$ different Supermasks. 
As a result, this can reduce the memory footprint since all Supermasks can be stored in the binary format.

% \vspace{-4mm}
\section{Experiments}
\label{sec:empirical_results}

\noindent\textbf{Model Architecture.}
For model architectures, we experiment with Transformer$_\text{small}$ and Transformer$_\text{base}$, following the same setting as in~\citet{Ott:2018scaling}: 6 encoder layers and 6 decoder layers on IWSLT14 and WMT14. 
We also vary the depth and width of the Transformer model on machine translation tasks. 
On IWSLT14, we use 3 different random seeds and plot the mean accuracy $\pm$ one standard deviation. 
All the embedding layers (including the final output projection layer) are also randomized and pruned unless otherwise specified. 
Moreover, on all figures, the ``fully-weighted model'' denotes the standard full model (all weights remaining). 

\noindent\textbf{Machine Translation results.}
In~\fref{fig:mt_base}, we present results for directly pruning a randomly weighted Transformer on IWSLT14 and WMT14 tasks. 
Specifically, we vary the ratio of remaining parameters in the randomized model.

As can be seen, there is no significant performance difference between a one-layer random Transformer versus a 6-layer standard random Transformer across different percents of remaining weights on IWSLT14 and WMT14. 
We also observe that having the remaining randomized weight percents approach 0 or 100 leads to the worst performance across the settings. 
This is expected since the outputs will be random when we have 100\% randomized weights, and the model will not perform well when only limited weights are unpruned (close to 0\%). 
The best performing subnetwork of a one-layer randomized Transformer has 50\% weights remained. 
Connected to the search space of the employed method where we are choosing $\sigma$\% out of 100\% randomized weights, $\sigma=50$ leads to the largest search space.

\noindent\textbf{Effectiveness of Pre-trained Embeddding layers.}
Embedding layers are critical since they can be viewed as the inputs for an NLP model, which are analogous to the image pixels in vision. 
Plenty of prior studies have explored how to obtain the pre-trained embedding in an un-supervised way~\citep{mikolov2013efficient,pennington-etal-2014-glove}. 
We experiment with this practical setting where we could have access to the encoder/decoder embedding layers, which are pre-trained from the public checkpoint in fairseq\footnote{\href{https://github.com/pytorch/fairseq/}{https://github.com/pytorch/fairseq/}}, and we present the results in ~\fref{fig:mt_embed}. 
We observe a significant performance boost for a one-layer randomized transformer across different remaining weights. 
The difference is much larger for the bigger WMT14 dataset (around +3.0 BLEU for WMT14 and +1.0 BLEU for IWSLT14). 
The best one-layer randomized Transformer reaches 89\%/74\% of the fully-weighted Transformer performance on IWSLT14/WMT14, respectively.

\begin{table}[]\resizebox{\linewidth}{!}{
% \small
\setlength\tabcolsep{1.5pt}
\begin{tabular}{c|ccccc}
\hline
\multicolumn{1}{c|}{Task} & Model & BLEU & \multicolumn{1}{c}{Memory} & \begin{tabular}[c]{@{}c@{}}Remaining \\ Param Ratio\end{tabular} & \begin{tabular}[c]{@{}c@{}}Param \\ (no mask)\end{tabular} \\\hline \midrule

\multirow{7}{*}{IWSLT} & Trans$_\text{small}$ & 34.66 ($\pm$0.11) & 148MB & 100.0 & 39M \\
\cline{2-6}

 & \begin{tabular}[c]{@{}c@{}}One-layer \\ Random Trans$_\text{small}$\end{tabular} & 30.95 ($\pm$0.12) & 28MB & 50.0 & 7M \\   \cline{2-6}
 & \begin{tabular}[c]{@{}c@{}}One-layer \\ Trans$_\text{wide}$\end{tabular} & 34.14 ($\pm$0.08) & 71MB & 50.0 & 18M \\\cline{2-6}
 & \begin{tabular}[c]{@{}c@{}}One-layer \\ Random Trans$_\text{deep}$\end{tabular} & 31.51 ($\pm$0.10) & 29MB & 50.0 & 7M \\\hline\midrule
\multirow{7}{*}{WMT} & Trans-base & 27.51 & 328MB & 100.0 & 86M \\\cline{2-6}
 & \begin{tabular}[c]{@{}c@{}}One-layer \\ Random Trans$_\text{base}$\end{tabular} & 20.35 & 96MB & 50.0 & 25M \\\cline{2-6}
 & \begin{tabular}[c]{@{}c@{}}One-layer \\ Random Trans$_\text{wider}$\end{tabular} & 25.24 & 227MB & 50.0 & 57M \\\cline{2-6}
 & \begin{tabular}[c]{@{}c@{}}One-layer\\ Random Trans$_\text{deeper}$\end{tabular} & 21.76 & 98MB & 50.0 & 25M\\\hline
\end{tabular}}
    \vspace{-5pt}
\caption{Machine Translation result for a fully-weighted Transformer versus one-layer random Transformer with pre-trained embedding layer (retain 50\% weights). IWSLT14 results are averaged over 3 random seeds, standard deviations are in brackets. }
\label{table:result}
    \vspace{-15pt}
\end{table}

\noindent\textbf{Effectiveness of Depth and Width.}
In~\tref{table:result}, we report the parameter size, BLEU score, and memory size of different one-layer randomized Transformers with  50\% remaining weights, where Trans$_\text{deep/deeper}$ are 12 encoder/decoder layers variant of Trans$_\text{small/base}$. Trans$_\text{wide/wider}$ have 2x hidden size as the Trans$_\text{small/base}$. 
The results are gathered with pre-trained encoder/decoder embedding~layers.\footnote{We use the checkpoint from FairSeq for Trans$_\text{base/big}$ on WMT14, and Trans$_\text{small}$ on IWSLT14 to obtain the pre-trained embedding layer for one-layer Trans$_\text{base/wider}$ and one-layer Trans$_\text{small}$. 
For one-layer Trans$_\text{wide}$ on IWSLT14, we pre-train fully-weighted model and then dump the embedding layer. 
Trans$_\text{deep/deeper}$ share the same embedding of the Trans$_\text{small/base}$. } 

Either increasing the depth or enlarging the width can improve the performance of our one-layer random transformer. 
Particularly, the deeper transformer can already achieve 79\%/90\% of the fully-weighted baseline models on WMT14/IWSLT14, respectively. 
For wider models, those numbers even increase to 92\%/98\%. 
This is mainly due to the larger search space introduced by the larger weight matrix. 
Another important point is that even when we increase/enlarge the depth/width of the model, the total memory consumption of these models is actually smaller than the standard baseline, since we only have one repeated layer and all the masks can be stored in a 1-bit setting. 

\begin{figure}
    \centering
    \includegraphics[width=0.8\linewidth]{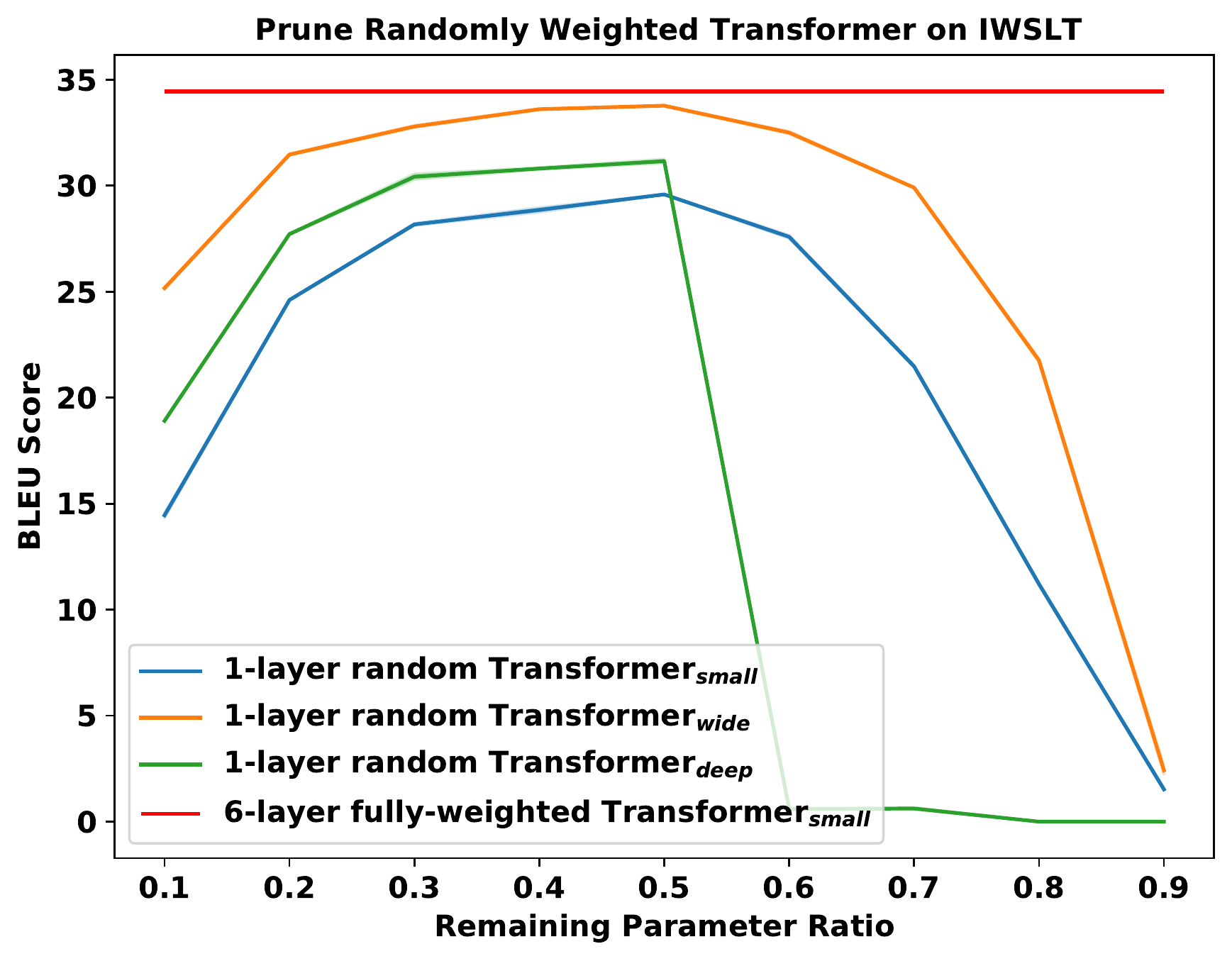}
        \vspace{-5pt}
    \caption{The effectiveness of depth and width.}
    \label{fig:res_iwslt_widedeep}
    \vspace{-10pt}
\end{figure}

Furthermore, we explore the effect of the different ratios of remaining parameters for different models on IWSLT14 in ~\fref{fig:res_iwslt_widedeep}. 
As can be seen, for the wider model, its performance is always better than the standard one across all different settings. 
However, for the deeper model, there is a sharp transition that happens at 50\%--60\% remaining parameters. 
The reason is that, given that our deeper model is twice as deep as the original, when we retain more random parameters ($>$50\%), the probability that the layer has a good ``subnetwork'' decreases significantly. 
This will lead the final probability to be $p_\text{smaller}^{2L}$ ($p_\text{smaller}<p$), which is much smaller than $p^L$ (see Section~\ref{sec:methodology}). 

\begin{figure}
    \centering
    \includegraphics[width=0.8\linewidth]{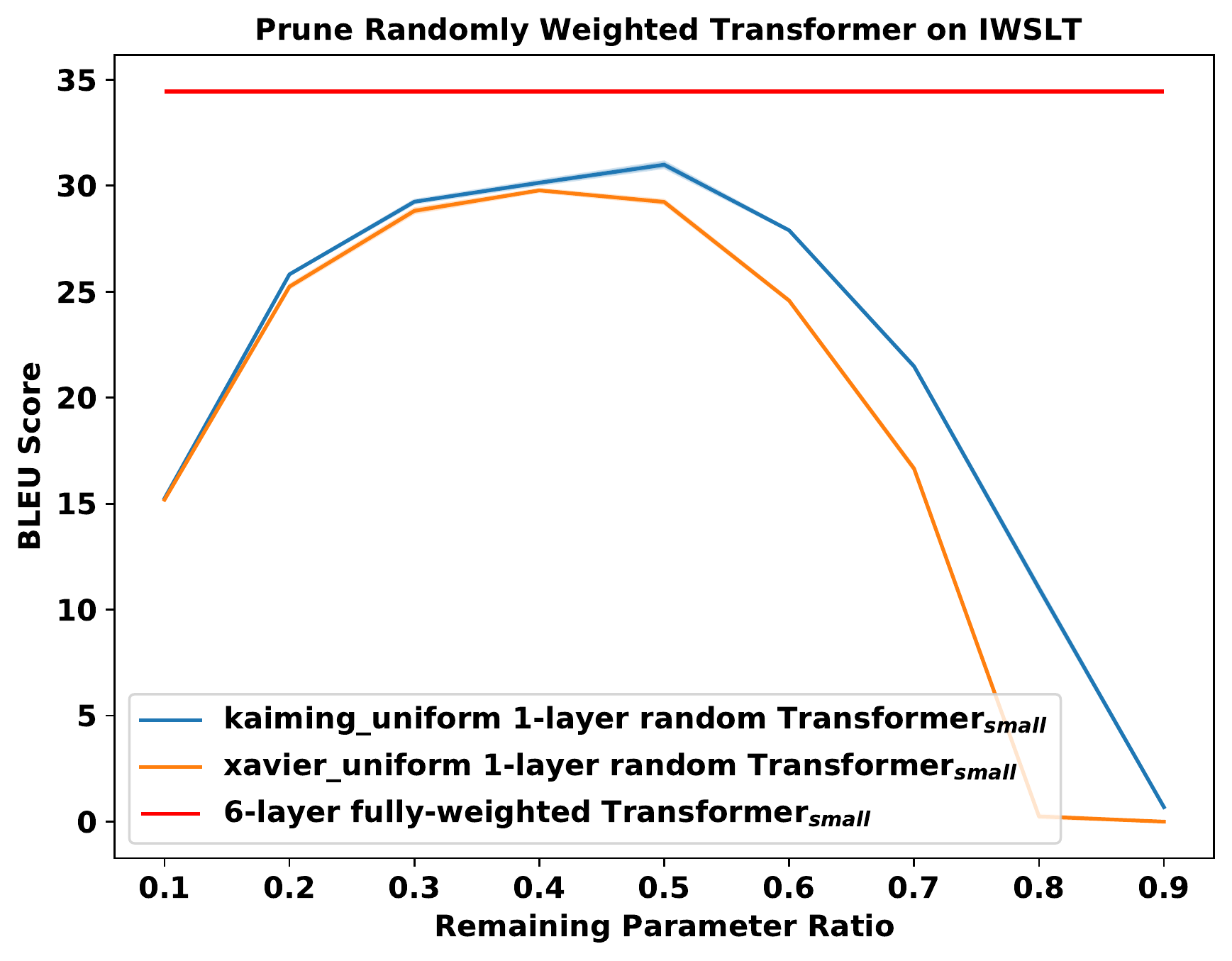}
            \vspace{-5pt}
    \caption{The effectiveness of different initialization.}
    \label{fig:res_iwslt_init}
    \vspace{-10pt}
\end{figure}

\noindent\textbf{Different Initialization.}
Weight initialization is one of the critical components to the success of the random feature~\citep{Wieting:2019notraining,Ramanujan:2020hidden,shen2020reservoir}. 
We experiment with kaiming uniform~\citep{Ramanujan:2020hidden} and Xavier uniform~\citep{Vaswani:2017attention} initialization methods, and we scale the standard deviation by $\sqrt{1/\sigma}$ when we retain $\sigma$ randomized weights. 
As shown in~\fref{fig:res_iwslt_init}, the performance of the one-layer randomized Transformer decreases when we switch to the Xavier uniform. 
The degradation becomes larger when more randomized weights retain in the network.

\begin{figure}
    \centering
    \includegraphics[width=0.8\linewidth]{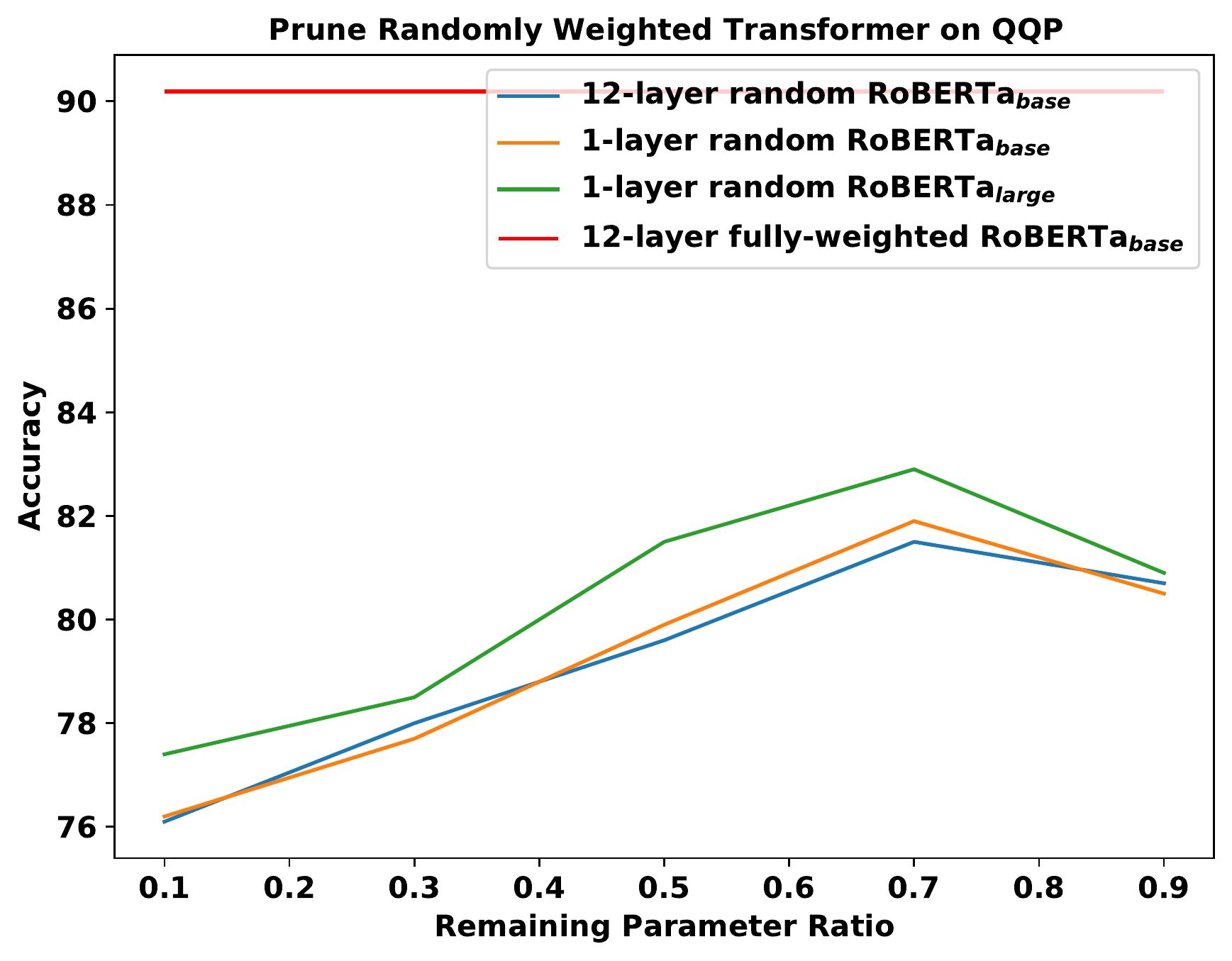}
    \caption{Prune Randomly Weighted Transformer performance on QQP .}
    \label{fig:qqp_base}
\end{figure}

\begin{figure}
    \centering
    \includegraphics[width=0.8\linewidth]{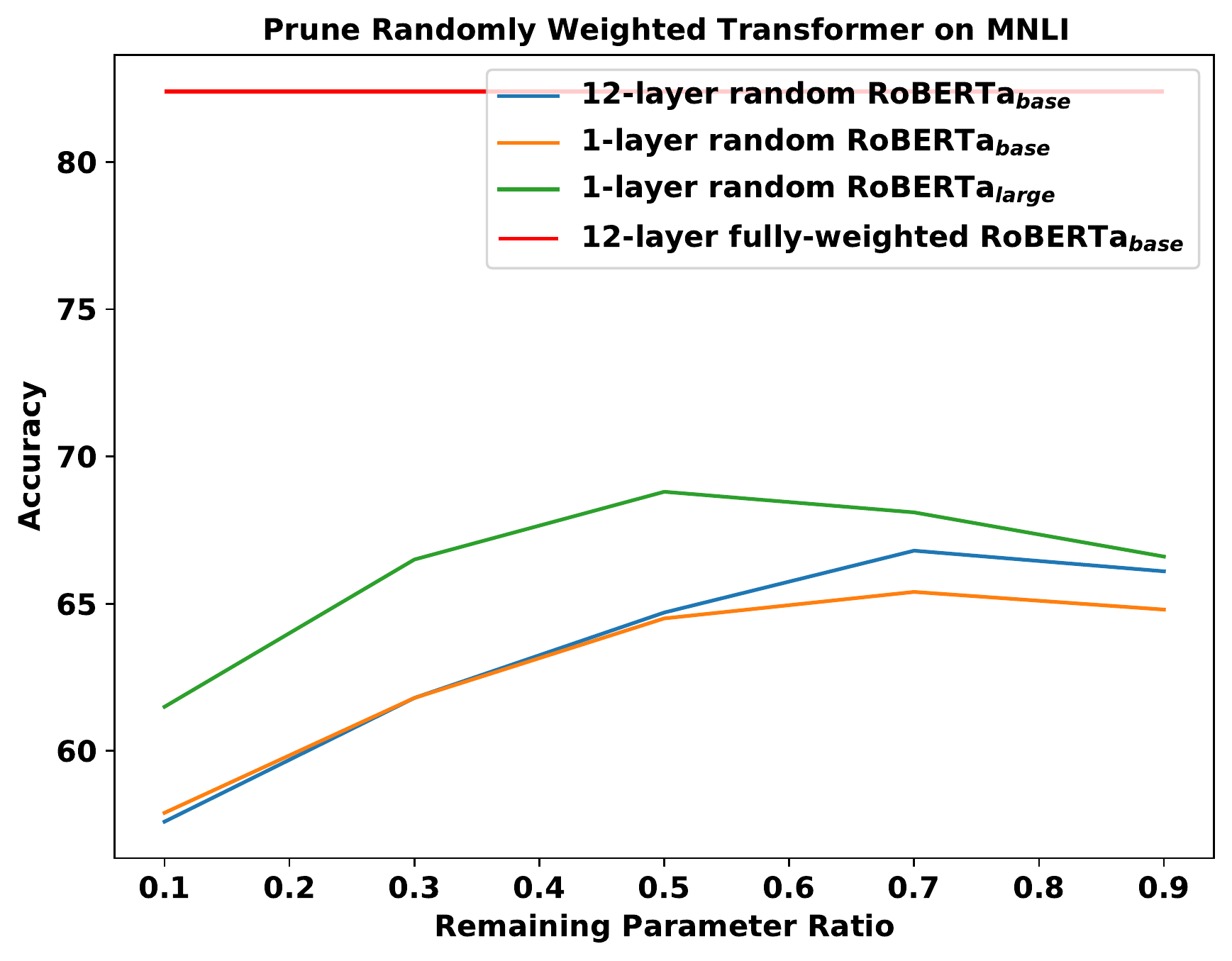}
    \caption{ Prune Randomly Weighted Transformer performance on MNLI.}
    \label{fig:mnli_base}
\end{figure}

\paragraph{QQP and MNLI results.} 
On QQP and MNLI, we experiment with RoBERTa$_\text{small}$ and RoBERTa$_\text{large}$, following \citet{Liu:2019roberta}. We use the pre-trained embedding layer of RoBERTa$_\text{base/large}$~\citep{Liu:2019roberta}. 
In~\fref{fig:qqp_base} and \ref{fig:mnli_base}, we show consistent results on QQP and MNLI, except that the best performing one-layer randomly weighted RoBERTa is achieved when we retain 70\% randomized weights, it reaches 79\%/91\% fully-weighted RoBERTa$_\text{base}$ accuracy on QQP and MNLI, respectively.  
The performance approaches 84\%/92\% of the aforementioned fully-weighted model performance when using the larger hidden size with one-layer randomly weighted RoBERTa$_\text{large}$. 

\paragraph{Implementation Details.} We evaluate on IWSLT14 de-en \citep{Cettolo:2015ReportOT} and WMT14 en-de \citep{bojar:2014-findings} for machine translation; QQP \citep{iyer2017qqp} and MultiNLI-matched (MNLI) \citep{Williams2017mnli} for natural language understanding.\footnote{For IWSLT, we follow the pre-processing steps in \citet{edunov:2018classical}. The train/val/test split is 129k/10k/6.8k sentences. 
For WMT, we follow pre-process as in \citet{Ott:2018scaling}, with 4.5M/16.5k/3k sentences in train/val/test.} 
We use 8 Volta V100 GPUs for WMT, and one V100 for IWSLT, QQP, and MNLI. The hyperparameters on IWSLT14 and WMT14 for training a one-layer randomized Transformer were set the same to the best-performing values from \citet{Ott:2018scaling} for training fully-weighted Transformer. The QQP and MNLI experiments followed \citet{Liu:2019roberta}.

% \vspace{-4mm}
\section{Conclusions}
\label{sec:conclusions}
In this paper, we validate the existence of effective subnetworks in a one-layer randomly weighted Transformer on translation tasks. 
Hidden within a one-layer randomly weighted Transformer$_\text{wide/wider}$ with fixed pre-trained embedding layers, we find there exist subnetworks that are smaller than, but can competitively match, the performance of a trained Transformer$_\text{small/base}$ on IWSLT14/WMT14.

% \section*{Ethical Considerations}
% One caveat of the proposed method is that data collected from public available resources that may contain biases, toxic contents, and other ethical issues. 
% This problem is common to AI models and we stress that de-biasing and a detailed examination are needed before deploying the system.
\section*{Acknowledgements}
We thank anonymous reviewers for their comments and suggestions. SS and KK were supported by grants from Samsung, Facebook, and the Berkeley Deep Drive Consortium. We would like to acknowledge DARPA, IARPA, NSF, and ONR for providing partial support of this work.
% \vspace{-4mm}

% \clearpage
\bibliography{ref}

\begin{thebibliography}{55}
\expandafter\ifx\csname natexlab\endcsname\relax\def\natexlab#1{#1}\fi

\bibitem[{Bai et~al.(2019)Bai, Kolter, and Koltun}]{bai2019deep}
Shaojie Bai, J~Zico Kolter, and Vladlen Koltun. 2019.
\newblock Deep equilibrium models.
\newblock \emph{Advances in Neural Information Processing Systems},
  32:690--701.

\bibitem[{Baum(1988)}]{Baum:1988jc}
Eric~B Baum. 1988.
\newblock On the capabilities of multilayer perceptrons.
\newblock \emph{Journal of complexity}, 4(3):193--215.

\bibitem[{Bengio et~al.(2013)Bengio, L{\'e}onard, and
  Courville}]{bengio2013estimating}
Yoshua Bengio, Nicholas L{\'e}onard, and Aaron Courville. 2013.
\newblock Estimating or propagating gradients through stochastic neurons for
  conditional computation.
\newblock \emph{arXiv preprint arXiv:1308.3432}.

\bibitem[{Bojar et~al.(2014)Bojar, Buck, Federmann, Haddow, Koehn, Leveling,
  Monz, Pecina, Post, Saint-Amand, Soricut, Specia, and
  Tamchyna}]{bojar:2014-findings}
Ond{\v{r}}ej Bojar, Christian Buck, Christian Federmann, Barry Haddow, Philipp
  Koehn, Johannes Leveling, Christof Monz, Pavel Pecina, Matt Post, Herve
  Saint-Amand, Radu Soricut, Lucia Specia, and Ale{\v{s}} Tamchyna. 2014.
\newblock Findings of the 2014 workshop on statistical machine translation.
\newblock In \emph{Proceedings of the Ninth Workshop on Statistical Machine
  Translation}, Baltimore, Maryland, USA. Association for Computational
  Linguistics.

\bibitem[{Brown et~al.(2020)Brown, Mann, Ryder, Subbiah, Kaplan, Dhariwal,
  Neelakantan, Shyam, Sastry, Askell et~al.}]{brown2020language}
Tom~B Brown, Benjamin Mann, Nick Ryder, Melanie Subbiah, Jared Kaplan, Prafulla
  Dhariwal, Arvind Neelakantan, Pranav Shyam, Girish Sastry, Amanda Askell,
  et~al. 2020.
\newblock Language models are few-shot learners.
\newblock \emph{arXiv preprint arXiv:2005.14165}.

\bibitem[{Cettolo et~al.(2015)Cettolo, Niehues, St{\"u}ker, Bentivogli, and
  Federico}]{Cettolo:2015ReportOT}
M.~Cettolo, J.~Niehues, S.~St{\"u}ker, L.~Bentivogli, and Marcello Federico.
  2015.
\newblock Report on the 11 th iwslt evaluation campaign , iwslt 2014.
\newblock In \emph{Proceedings of IWSLT}.

\bibitem[{Chen et~al.(2020)Chen, Frankle, Chang, Liu, Zhang, Wang, and
  Carbin}]{chen2020lottery}
Tianlong Chen, Jonathan Frankle, Shiyu Chang, Sijia Liu, Yang Zhang, Zhangyang
  Wang, and Michael Carbin. 2020.
\newblock The lottery ticket hypothesis for pre-trained bert networks.
\newblock \emph{arXiv preprint arXiv:2007.12223}.

\bibitem[{Choromanski et~al.(2020)Choromanski, Likhosherstov, Dohan, Song,
  Davis, Sarlos, Belanger, Colwell, and Weller}]{Choromanski:2020performer}
Krzysztof Choromanski, Valerii Likhosherstov, David Dohan, Xingyou Song, Jared
  Davis, Tamas Sarlos, David Belanger, Lucy Colwell, and Adrian Weller. 2020.
\newblock Masked language modeling for proteins via linearly scalable
  long-context transformers.
\newblock \emph{arXiv preprint arXiv:2006.03555}.

\bibitem[{Conneau et~al.(2017)Conneau, Kiela, Schwenk, Barrault, and
  Bordes}]{Conneau:2017infersent}
Alexis Conneau, Douwe Kiela, Holger Schwenk, Loic Barrault, and Antoine Bordes.
  2017.
\newblock Supervised learning of universal sentence representations from
  natural language inference data.
\newblock \emph{arXiv preprint arXiv:1705.02364}.

\bibitem[{Dehghani et~al.(2018)Dehghani, Gouws, Vinyals, Uszkoreit, and
  Kaiser}]{dehghani2018universal}
Mostafa Dehghani, Stephan Gouws, Oriol Vinyals, Jakob Uszkoreit, and Lukasz
  Kaiser. 2018.
\newblock Universal transformers.
\newblock In \emph{International Conference on Learning Representations}.

\bibitem[{Devlin et~al.(2018)Devlin, Chang, Lee, and
  Toutanova}]{Devlin:2018bert}
Jacob Devlin, Ming-Wei Chang, Kenton Lee, and Kristina Toutanova. 2018.
\newblock Bert: Pre-training of deep bidirectional transformers for language
  understanding.
\newblock \emph{arXiv preprint arXiv:1810.04805}.

\bibitem[{Edunov et~al.(2018)Edunov, Ott, Auli, Grangier, and
  Ranzato}]{edunov:2018classical}
Sergey Edunov, Myle Ott, Michael Auli, David Grangier, and Marc{'}Aurelio
  Ranzato. 2018.
\newblock Classical structured prediction losses for sequence to sequence
  learning.
\newblock In \emph{Proceedings of the 2018 Conference of the North {A}merican
  Chapter of the Association for Computational Linguistics: Human Language
  Technologies, Volume 1 (Long Papers)}, New Orleans, Louisiana. Association
  for Computational Linguistics.

\bibitem[{Fan et~al.(2019)Fan, Grave, and Joulin}]{fan2019reducing}
Angela Fan, Edouard Grave, and Armand Joulin. 2019.
\newblock Reducing transformer depth on demand with structured dropout.
\newblock In \emph{International Conference on Learning Representations}.

\bibitem[{Frankle and Carbin(2018)}]{frankle2018lottery}
Jonathan Frankle and Michael Carbin. 2018.
\newblock The lottery ticket hypothesis: Finding sparse, trainable neural
  networks.
\newblock \emph{arXiv preprint arXiv:1803.03635}.

\bibitem[{Frankle et~al.(2020)Frankle, Dziugaite, Roy, and
  Carbin}]{frankle2020linear}
Jonathan Frankle, Gintare~Karolina Dziugaite, Daniel Roy, and Michael Carbin.
  2020.
\newblock Linear mode connectivity and the lottery ticket hypothesis.
\newblock In \emph{International Conference on Machine Learning}, pages
  3259--3269. PMLR.

\bibitem[{Gaier and Ha(2019)}]{gaier2019weight}
Adam Gaier and David Ha. 2019.
\newblock Weight agnostic neural networks.
\newblock \emph{arXiv preprint arXiv:1906.04358}.

\bibitem[{Gallicchio and Micheli(2017)}]{Gallicchio:2017echo}
Claudio Gallicchio and Alessio Micheli. 2017.
\newblock Echo state property of deep reservoir computing networks.
\newblock \emph{Cognitive Computation}, 9(3):337--350.

\bibitem[{Gamba et~al.(1961)Gamba, Gamberini, Palmieri, and
  Sanna}]{Gamba:1961papa}
A.~Gamba, L.~Gamberini, G.~Palmieri, and R.~Sanna. 1961.
\newblock Further experiments with papa.
\newblock \emph{Il Nuovo Cimento (1955-1965)}, 20(2):112--115.

\bibitem[{Gan et~al.(2021)Gan, Chen, Li, Chen, Cheng, Wang, and
  Liu}]{gan2021playing}
Zhe Gan, Yen-Chun Chen, Linjie Li, Tianlong Chen, Yu~Cheng, Shuohang Wang, and
  Jingjing Liu. 2021.
\newblock Playing lottery tickets with vision and language.
\newblock \emph{arXiv preprint arXiv:2104.11832}.

\bibitem[{Iyer et~al.(2017)Iyer, Dandekar, and Csernai}]{iyer2017qqp}
Shankar Iyer, Nikhil Dandekar, and Kornl Csernai. 2017.
\newblock First quora dataset release: Question pairs, 2017.
\newblock \emph{URL https://data. quora.
  com/First-Quora-Dataset-Release-Question-Pairs}.

\bibitem[{Jaeger(2003)}]{Jaeger:2003echostate}
Herbert Jaeger. 2003.
\newblock Adaptive nonlinear system identification with echo state networks.
\newblock In \emph{Advances in neural information processing systems}.

\bibitem[{Jiao et~al.(2020)Jiao, Yin, Shang, Jiang, Chen, Li, Wang, and
  Liu}]{jiao2020tinybert}
Xiaoqi Jiao, Yichun Yin, Lifeng Shang, Xin Jiang, Xiao Chen, Linlin Li, Fang
  Wang, and Qun Liu. 2020.
\newblock Tinybert: Distilling bert for natural language understanding.
\newblock In \emph{Proceedings of the 2020 Conference on Empirical Methods in
  Natural Language Processing: Findings}, pages 4163--4174.

\bibitem[{Lan et~al.(2020)Lan, Chen, Goodman, Gimpel, Sharma, and
  Soricut}]{lan2020albert}
Zhenzhong Lan, Mingda Chen, Sebastian Goodman, Kevin Gimpel, Piyush Sharma, and
  Radu Soricut. 2020.
\newblock Albert: A lite bert for self-supervised learning of language
  representations.

\bibitem[{Li et~al.(2020)Li, Wallace, Shen, Lin, Keutzer, Klein, and
  Gonzalez}]{li2020train}
Zhuohan Li, Eric Wallace, Sheng Shen, Kevin Lin, Kurt Keutzer, Dan Klein, and
  Joey Gonzalez. 2020.
\newblock Train big, then compress: Rethinking model size for efficient
  training and inference of transformers.
\newblock In \emph{International Conference on Machine Learning}. PMLR.

\bibitem[{Liu et~al.(2019)Liu, Ott, Goyal, Du, Joshi, Chen, Levy, Lewis,
  Zettlemoyer, and Stoyanov}]{Liu:2019roberta}
Yinhan Liu, Myle Ott, Naman Goyal, Jingfei Du, Mandar Joshi, Danqi Chen, Omer
  Levy, Mike Lewis, Luke Zettlemoyer, and Veselin Stoyanov. 2019.
\newblock Roberta: A robustly optimized bert pretraining approach.
\newblock \emph{arXiv preprint arXiv:1907.11692}.

\bibitem[{Luko{\v{s}}evi{\v{c}}ius and
  Jaeger(2009)}]{Lukovsevivcius:2009reservoir}
Mantas Luko{\v{s}}evi{\v{c}}ius and Herbert Jaeger. 2009.
\newblock Reservoir computing approaches to recurrent neural network training.
\newblock \emph{Computer Science Review}, 3(3).

\bibitem[{Maass et~al.(2002)Maass, Natschl{\"a}ger, and
  Markram}]{Maass:2002lsm}
Wolfgang Maass, Thomas Natschl{\"a}ger, and Henry Markram. 2002.
\newblock Real-time computing without stable states: A new framework for neural
  computation based on perturbations.
\newblock \emph{Neural computation}, 14(11):2531--2560.

\bibitem[{Michel et~al.(2019)Michel, Levy, and Neubig}]{michel2019sixteen}
Paul Michel, Omer Levy, and Graham Neubig. 2019.
\newblock Are sixteen heads really better than one?
\newblock \emph{Advances in Neural Information Processing Systems},
  32:14014--14024.

\bibitem[{Mikolov et~al.(2013)Mikolov, Chen, Corrado, and
  Dean}]{mikolov2013efficient}
Tomas Mikolov, Kai Chen, Greg Corrado, and Jeffrey Dean. 2013.
\newblock Efficient estimation of word representations in vector space.
\newblock \emph{arXiv preprint arXiv:1301.3781}.

\bibitem[{Ott et~al.(2018)Ott, Edunov, Grangier, and Auli}]{Ott:2018scaling}
Myle Ott, Sergey Edunov, David Grangier, and Michael Auli. 2018.
\newblock Scaling neural machine translation.
\newblock \emph{arXiv preprint arXiv:1806.00187}.

\bibitem[{Pao et~al.(1994)Pao, Park, and Sobajic}]{Pao:1994nc}
Yoh-Han Pao, Gwang-Hoon Park, and Dejan~J Sobajic. 1994.
\newblock Learning and generalization characteristics of the random vector
  functional-link net.
\newblock \emph{Neurocomputing}, 6(2):163--180.

\bibitem[{Peng et~al.(2021)Peng, Pappas, Yogatama, Schwartz, Smith, and
  Kong}]{peng2021random}
Hao Peng, Nikolaos Pappas, Dani Yogatama, Roy Schwartz, Noah Smith, and
  Lingpeng Kong. 2021.
\newblock Random feature attention.
\newblock In \emph{International Conference on Learning Representations}.

\bibitem[{Pennington et~al.(2014)Pennington, Socher, and
  Manning}]{pennington-etal-2014-glove}
Jeffrey Pennington, Richard Socher, and Christopher Manning. 2014.
\newblock \href {https://doi.org/10.3115/v1/D14-1162} {{G}lo{V}e: Global
  vectors for word representation}.
\newblock In \emph{Proceedings of the 2014 Conference on Empirical Methods in
  Natural Language Processing ({EMNLP})}, pages 1532--1543, Doha, Qatar.
  Association for Computational Linguistics.

\bibitem[{Pilault et~al.(2020)Pilault, Park, and Pal}]{Pilault:2020impressive}
Jonathan Pilault, Jaehong Park, and Christopher Pal. 2020.
\newblock On the impressive performance of randomly weighted encoders in
  summarization tasks.
\newblock \emph{arXiv preprint arXiv:2002.09084}.

\bibitem[{Prasanna et~al.(2020)Prasanna, Rogers, and
  Rumshisky}]{prasanna2020bert}
Sai Prasanna, Anna Rogers, and Anna Rumshisky. 2020.
\newblock When bert plays the lottery, all tickets are winning.
\newblock \emph{arXiv preprint arXiv:2005.00561}.

\bibitem[{Raffel et~al.(2019)Raffel, Shazeer, Roberts, Lee, Narang, Matena,
  Zhou, Li, and Liu}]{raffel2019exploring}
Colin Raffel, Noam Shazeer, Adam Roberts, Katherine Lee, Sharan Narang, Michael
  Matena, Yanqi Zhou, Wei Li, and Peter~J Liu. 2019.
\newblock Exploring the limits of transfer learning with a unified text-to-text
  transformer.
\newblock \emph{arXiv preprint arXiv:1910.10683}.

\bibitem[{Rahimi and Recht(2008)}]{Rahimi:2008random}
Ali Rahimi and Benjamin Recht. 2008.
\newblock Random features for large-scale kernel machines.
\newblock In \emph{Advances in neural information processing systems}, pages
  1177--1184.

\bibitem[{Rahimi and Recht(2009)}]{Rahimi:2009kitchen}
Ali Rahimi and Benjamin Recht. 2009.
\newblock Weighted sums of random kitchen sinks: Replacing minimization with
  randomization in learning.
\newblock In \emph{Advances in neural information processing systems}, pages
  1313--1320.

\bibitem[{Ramanujan et~al.(2020)Ramanujan, Wortsman, Kembhavi, Farhadi, and
  Rastegari}]{Ramanujan:2020hidden}
Vivek Ramanujan, Mitchell Wortsman, Aniruddha Kembhavi, Ali Farhadi, and
  Mohammad Rastegari. 2020.
\newblock What's hidden in a randomly weighted neural network?
\newblock In \emph{Proceedings of the IEEE/CVF Conference on Computer Vision
  and Pattern Recognition}, pages 11893--11902.

\bibitem[{Renda et~al.(2020)Renda, Frankle, and Carbin}]{renda2020comparing}
Alex Renda, Jonathan Frankle, and Michael Carbin. 2020.
\newblock Comparing rewinding and fine-tuning in neural network pruning.
\newblock \emph{arXiv preprint arXiv:2003.02389}.

\bibitem[{Sanh et~al.(2020)Sanh, Wolf, and Rush}]{sanh2020movement}
Victor Sanh, Thomas Wolf, and Alexander Rush. 2020.
\newblock Movement pruning: Adaptive sparsity by fine-tuning.
\newblock \emph{Advances in Neural Information Processing Systems}, 33.

\bibitem[{Scardapane and Wang(2017)}]{Scardapane:2017randomness}
Simone Scardapane and Dianhui Wang. 2017.
\newblock Randomness in neural networks: an overview.
\newblock \emph{Wiley Interdisciplinary Reviews: Data Mining and Knowledge
  Discovery}, 7(2):e1200.

\bibitem[{Schmidt et~al.(1992)Schmidt, Kraaijveld, and Duin}]{Schmidt:1992pr}
Wouter~F Schmidt, Martin~A Kraaijveld, and Robert~PW Duin. 1992.
\newblock Feedforward neural networks with random weights.
\newblock In \emph{Proceedings of the 11th International Conference on Pattern
  Recognition, 1992. Vol. II. Conference B: Pattern Recognition Methodology and
  Systems}, pages 1--4.

\bibitem[{Shen et~al.(2021)Shen, Baevski, Morcos, Keutzer, Auli, and
  Kiela}]{shen2020reservoir}
Sheng Shen, Alexei Baevski, Ari~S Morcos, Kurt Keutzer, Michael Auli, and Douwe
  Kiela. 2021.
\newblock Reservoir transformers.
\newblock In \emph{ACL}.

\bibitem[{Shen et~al.(2020)Shen, Dong, Ye, Ma, Yao, Gholami, Mahoney, and
  Keutzer}]{shen2020q}
Sheng Shen, Zhen Dong, Jiayu Ye, Linjian Ma, Zhewei Yao, Amir Gholami,
  Michael~W Mahoney, and Kurt Keutzer. 2020.
\newblock Q-bert: Hessian based ultra low precision quantization of bert.
\newblock In \emph{Proceedings of the AAAI Conference on Artificial
  Intelligence}, volume~34, pages 8815--8821.

\bibitem[{Shoeybi et~al.(2019)Shoeybi, Patwary, Puri, LeGresley, Casper, and
  Catanzaro}]{shoeybi2019megatron}
Mohammad Shoeybi, Mostofa Patwary, Raul Puri, Patrick LeGresley, Jared Casper,
  and Bryan Catanzaro. 2019.
\newblock {Megatron-LM}: Training multi-billion parameter language models using
  gpu model parallelism.
\newblock \emph{arXiv preprint arXiv:1909.08053}.

\bibitem[{Sun et~al.(2020)Sun, Yu, Song, Liu, Yang, and
  Zhou}]{sun2020mobilebert}
Zhiqing Sun, Hongkun Yu, Xiaodan Song, Renjie Liu, Yiming Yang, and Denny Zhou.
  2020.
\newblock Mobilebert: a compact task-agnostic bert for resource-limited
  devices.
\newblock In \emph{Proceedings of the 58th Annual Meeting of the Association
  for Computational Linguistics}, pages 2158--2170.

\bibitem[{Vaswani et~al.(2017)Vaswani, Shazeer, Parmar, Uszkoreit, Jones,
  Gomez, Kaiser, and Polosukhin}]{Vaswani:2017attention}
Ashish Vaswani, Noam Shazeer, Niki Parmar, Jakob Uszkoreit, Llion Jones,
  Aidan~N Gomez, {\L}ukasz Kaiser, and Illia Polosukhin. 2017.
\newblock Attention is all you need.
\newblock In \emph{Advances in neural information processing systems}, pages
  5998--6008.

\bibitem[{Voita and Titov(2020)}]{voita2020information}
Elena Voita and Ivan Titov. 2020.
\newblock Information-theoretic probing with minimum description length.
\newblock \emph{arXiv preprint arXiv:2003.12298}.

\bibitem[{Wieting and Kiela(2019)}]{Wieting:2019notraining}
John Wieting and Douwe Kiela. 2019.
\newblock No training required: Exploring random encoders for sentence
  classification.
\newblock \emph{arXiv preprint arXiv:1901.10444}.

\bibitem[{Williams et~al.(2017)Williams, Nangia, and Bowman}]{Williams2017mnli}
Adina Williams, Nikita Nangia, and Samuel~R Bowman. 2017.
\newblock A broad-coverage challenge corpus for sentence understanding through
  inference.
\newblock \emph{arXiv preprint arXiv:1704.05426}.

\bibitem[{Wortsman et~al.(2020)Wortsman, Ramanujan, Liu, Kembhavi, Rastegari,
  Yosinski, and Farhadi}]{wortsman2020supermasks}
Mitchell Wortsman, Vivek Ramanujan, Rosanne Liu, Aniruddha Kembhavi, Mohammad
  Rastegari, Jason Yosinski, and Ali Farhadi. 2020.
\newblock Supermasks in superposition for continual learning.
\newblock \emph{Advances in Neural Information Processing Systems (NeurIPS)},
  6.

\bibitem[{Yao et~al.(2021)Yao, Ma, Shen, Keutzer, and
  Mahoney}]{yao2021mlpruning}
Zhewei Yao, Linjian Ma, Sheng Shen, Kurt Keutzer, and Michael~W Mahoney. 2021.
\newblock Mlpruning: A multilevel structured pruning framework for
  transformer-based models.
\newblock \emph{arXiv preprint arXiv:2105.14636}.

\bibitem[{Yu et~al.(2019)Yu, Edunov, Tian, and Morcos}]{yu2019playing}
Haonan Yu, Sergey Edunov, Yuandong Tian, and Ari~S Morcos. 2019.
\newblock Playing the lottery with rewards and multiple languages: lottery
  tickets in rl and nlp.
\newblock \emph{arXiv preprint arXiv:1906.02768}.

\bibitem[{Zhou et~al.(2019)Zhou, Lan, Liu, and
  Yosinski}]{Zhou:2019deconstructing}
Hattie Zhou, Janice Lan, Rosanne Liu, and Jason Yosinski. 2019.
\newblock Deconstructing lottery tickets: Zeros, signs, and the supermask.
\newblock In \emph{Advances in Neural Information Processing Systems}, pages
  3597--3607.

\end{thebibliography}
\bibliographystyle{acl_natbib}

\clearpage

\end{document}

\maketitle

% %%%%%%%% BODY TEXT
% \input _s0_abstract.tex
% % \vspace{-4mm}
% \input _s1_intro.tex
% % \vspace{-4mm}
% \input _s2_related_work.tex
% \input _s3_method.tex
% % \vspace{-4mm}
% \input _s4_results.tex
% % \vspace{-4mm}
% \input _s5_discussion.tex
% % \vspace{-4mm}
% \input _s6_conclusion.tex
% % \vspace{-4mm}

\input _s7_appendix.tex
% \clearpage
\bibliography{ref}
\bibliographystyle{acl_natbib}

\clearpage